\documentclass[11pt,a4paper]{article}
\usepackage[utf8]{inputenc}
\usepackage{authblk}
\usepackage{amsmath}
\usepackage{makecell} 
\usepackage{amsfonts}
\usepackage{booktabs}
\usepackage{hyperref}
\usepackage{geometry}
\usepackage{tabularx}
\usepackage{tcolorbox}
\usepackage{svg}
\usepackage{graphicx}
\usepackage{subcaption}
\usepackage{lipsum}
\usepackage{cite}
\usepackage{amsmath}
\usepackage{algorithmic}
\usepackage{array}
\usepackage{hyperref}
\usepackage{orcidlink}
\usepackage{stfloats}
\usepackage{tcolorbox}
\usepackage{verbatim}   
\tcbuselibrary{listings, breakable} 
\usepackage{listings}           
\usepackage{colortbl}   
\usepackage{pifont}     
\usepackage{tabularx}

\title{\textbf{Beyond Literal Summarization: Redefining Hallucination for Medical SOAP Note Evaluation}}

\author{Augnito Research\thanks{
\textbf{Contributors:} Bhavik Vachhani, Kush Shrisvastava, Pranshu Nema, Sai Chiranthan.\\
\textbf{Acknowledgments:} The authors thank Dr. Nikhil Saldanha for manual verification of the clinical notes.
}}
\date{April 2026}

\begin{document}

\maketitle

\begin{abstract}
Evaluating large language models (LLMs) for clinical documentation tasks such as SOAP note generation remains challenging. Unlike standard summarization, these tasks require clinical abstraction, normalization of colloquial language, and medically grounded inference. However, prevailing evaluation methods including automated metrics and LLM as judge frameworks rely on lexical faithfulness, often labeling any information not explicitly present in the transcript as hallucination.

We show that such approaches systematically misclassify clinically valid outputs as errors, inflating hallucination rates and distorting model assessment. Our analysis reveals that many flagged hallucinations correspond to legitimate clinical transformations, including synonym mapping, abstraction of examination findings, diagnostic inference, and guideline consistent care planning.

By aligning evaluation criteria with clinical reasoning through calibrated prompting and retrieval grounded in medical ontologies we observe a significant shift in outcomes. Under a lexical evaluation regime, the mean hallucination rate is 35\%, heavily penalizing valid reasoning. With inference aware evaluation, this drops to 9\%, with remaining cases reflecting genuine safety concerns. These findings suggest that current evaluation practices over penalize valid clinical reasoning and may measure artifacts of evaluation design rather than true errors, underscoring the need for clinically informed evaluation in high context domains like medicine.

\end{abstract}

\section{Introduction}
The integration of artificial intelligence into clinical workflows has accelerated dramatically in recent years, driven by the dual pressures of physician burnout and the administrative burden of clinical documentation. It is estimated that physicians spend nearly two hours on electronic health record (EHR) documentation for every one hour of direct patient care \cite{Sinsky2016}, contributing to reduced patient interaction quality and systemic inefficiency. Large language models (LLMs), trained on vast corpora of biomedical and general text, have emerged as compelling candidates for automating this burden. Among the most widely adopted clinical document types is the SOAP note, a structured format capturing the Subjective experience of the patient, Objective clinical findings, the clinician's Assessment, and the Plan for treatment. LLMs have demonstrated the ability to generate coherent, well structured SOAP notes from raw physician–patient conversations, substantially reducing documentation time.At first glance, this task appears similar to summarization. However, clinical documentation is fundamentally an act of interpretation rather than transcription. A high quality SOAP note does not merely restate what was said, it normalizes colloquial expressions into medical terminology, structures findings according to clinical conventions, and infers diagnoses or care plans from incomplete or implicit evidence. Correctness, therefore, depends on clinical validity rather than strict textual overlap.This distinction introduces a critical challenge for evaluation.

Traditional notions of faithfulness, whether based on lexical overlap or strict grounding to the source text, fail to capture the interpretive nature of clinical documentation. In practice, many evaluation frameworks implicitly adopt what we term over literal evaluation bias: the tendency to equate the absence of explicit textual evidence with lack of validity. Under this bias, any information not directly traceable to the transcript is treated as a hallucination. Hallucination, broadly defined as the generation of unsupported or fabricated content, is a well recognized failure mode in LLMs. In clinical settings, its consequences can be severe. A hallucinated drug allergy, a fabricated comorbidity, or an unsupported diagnosis can propagate through the medical record and lead to inappropriate or even dangerous care decisions. Yet the medical domain introduces a crucial nuance that generic hallucination frameworks fail to capture: clinical inference. A physician does not merely transcribe what a patient says, they interpret, synthesize, and reason. Consider a patient presenting with postprandial epigastric burning, retrosternal radiation, nocturnal pain awakening from sleep, and six months of daily ibuprofen use without food. The clinician may never explicitly state “peptic ulcer disease” or “GERD” yet documenting these as working diagnoses in the Assessment is both standard and necessary. Under a strictly literal evaluation framework, such diagnoses would be flagged as hallucinations. Under a clinically informed framework, they represent appropriate medical reasoning. This tension has direct consequences for how we evaluate and deploy LLMs in healthcare. An evaluation framework that over penalizes valid inference will suppress the very reasoning capabilities that make clinical documentation useful. A structural source of this imbalance lies in how evaluation is framed. Many approaches implicitly treat correctness as requiring explicit textual grounding, checking whether each piece of generated content can be directly traced to the source transcript. In clinical settings, this assumption breaks down. Documentation routinely involves paraphrasing, normalization, and inference, which may be clinically valid without a one to one textual match. When evaluated under strictly literal criteria, such transformations are systematically misclassified as hallucinations. We refer to this as over literal evaluation bias: the tendency to equate absence of explicit evidence with lack of validity. As a result, evaluators default to conservative judgments under uncertainty, leading to inflated hallucination rates \cite{kalai2025languagemodelshallucinate}.Conversely, a framework that is too permissive risks overlooking genuinely dangerous fabrications. Striking this balance requires rethinking not just models, but evaluation itself.

\section{Background and Related Work}

\subsection{Hallucination in Large Language Models}
Hallucination in neural text generation was formally characterized by \cite{Maynez2020} in the context of abstractive summarization, where models were observed to generate content that was either intrinsic (contradicting the source) or extrinsic (introducing information not present in the source). Subsequent work by \cite{Ji2023} provided a comprehensive taxonomy of hallucination across tasks and model architectures, noting that hallucination tends to increase in tasks requiring long form generation or domain specific knowledge retrieval both hallmarks of clinical documentation.
With the advent of instruction tuned LLMs hallucination patterns have become more subtle and difficult to detect. Rather than outright fabrications, modern models tend to produce plausible sounding but unverifiable content, a phenomenon sometimes called confident confabulation \cite{Huang2023}.  This is particularly dangerous in the medical domain, where clinical plausibility and factual accuracy are not equivalent.
\subsection{Hallucination in Clinical NLP}
The clinical NLP community has long grappled with reliability and faithfulness in automated text generation. Early work on visit note generation in \cite{Yim2023} noted that neural models frequently introduced diagnoses not supported by by the source dialogue. More recent studies examining LLM generated clinical documentation have raised similar alarms. Authors in \cite{VanVeen2023} evaluated LLMs on clinical note generation tasks and found hallucination rates ranging from 15–35\% depending on model and task complexity. Authors here \cite{Umapathi2023} demonstrated that even state of the art models struggled to maintain factual fidelity in medical question answering, often generating fabricated drug dosages or clinical guidelines. Critically, these studies typically employed strict factual grounding criteria, without accounting for the inference layer inherent to clinical reasoning. Med PaLM\cite{Singhal2023} introduced, a medically fine tuned LLM evaluated against physician curated benchmarks. Their evaluation rubric acknowledged that model responses should be assessed for both factual accuracy and clinical reasoning quality, implicitly recognizing that medical inference is not the same as fabrication,  a distinction this paper formalizes.

\subsection{Clinical Documentation is Not Summarization}

At a surface level, generating a SOAP note from a doctor patient conversation resembles summarization \cite{Krishna2021}. Both involve condensing information into a structured output. However, this analogy breaks down in clinical settings, where correctness depends not only on 
what is said, but on how it is interpreted.

Clinical documentation involves multiple layers of transformation. Patient language is often colloquial and imprecise, requiring normalization into standardized medical terminology. Clinically relevant information may be implicit, requiring inference based on symptom patterns, duration, and history. Documentation also follows conventions, where normal findings are summarized using standard phrases even if not 
explicitly stated \cite{Gao2022}.

These transformations can be understood as clinically grounded abstractions. They introduce information that may not appear verbatim in the source but is nonetheless correct and necessary. This stands in contrast to traditional summarization, where fidelity is defined in 
terms of textual overlap. As a result, evaluating clinical documentation using summarization based criteria introduces a fundamental mismatch. Models that perform appropriate clinical reasoning may appear to deviate from the source, while models that adhere closely to surface form may miss clinically important insights.

\subsection{Limitations of Existing Evaluation Metrics}

Traditional metrics such as ROUGE and BLEU measure lexical overlap and are therefore poorly suited for clinical tasks. Multiple clinically valid outputs may differ significantly in wording while conveying the same meaning. For example, “burning chest sensation” may be documented as “heartburn” or interpreted as “gastroesophageal reflux disease”.

To address these limitations, evaluation has shifted toward model based approaches, particularly LLM as Judge frameworks. While more flexible, these systems often inherit the same limitations when guided by overly strict notions of grounding. Even when capable of semantic reasoning, they may penalize valid outputs if instructed to prioritize explicit textual evidence.

This suggests that the core challenge lies not in the choice of evaluator, but in how correctness is defined. In domains like medicine, where abstraction and reasoning are integral, evaluation must move beyond surface level faithfulness.

\subsection{LLM as Judge Evaluation}
The LLM as Judge paradigm, formalized by \cite{Zheng2023} in the context of general purpose model evaluation, proposes using a capable LLM (typically GPT 4) as an evaluator of other model outputs. This approach has shown strong correlation with human expert judgments in several domains, and has been adapted for medical evaluation tasks. However, the validity of LLM as Judge hinges critically on the quality of the evaluation prompt. A judge prompt that lacks domain calibration,  for instance, one that treats all unanchored claims as hallucinations without regard for clinical reasoning will systematically over flag legitimate medical inference as hallucination, producing artificially inflated error rates. Designing inference aware judge prompts is therefore a non trivial and consequential challenge in clinical NLP evaluation. Beyond prompt design, the effectiveness of a clinical judge also depends on the breadth of domain knowledge available to it during evaluation including terminology mappings, diagnostic ontologies, and pharmacological equivalences, knowledge that may exceed what is reliably encoded in a model's parametric weights alone.

\subsection{From Human Evaluation to LLM Judges}

Human evaluation remains the gold standard for assessing clinical documentation, capturing nuanced aspects of correctness, completeness, and safety. However, it is difficult to scale and subject to variability across evaluators.

LLM based evaluation offers a scalable alternative, but introduces its own challenges. The apparent disagreement between human experts and LLM judges often reflects differences in evaluation criteria rather than true differences in output quality. Clinicians readily accept paraphrasing, abstraction, and inference, whereas LLM judges constrained by strict grounding may not.

These observations underscore a key point: the effectiveness of LLM as Judge depends not only on model capability, but on the evaluation policy encoded in its instructions. Aligning this policy with clinical reasoning is essential for meaningful evaluation.

\section{Defining Hallucination in Clinical SOAP Note Generation}
\label{sec:section_3}
\subsection{The Source Claim Framework}
We define the SOAP note generation task as follows: given a physician patient transcript T, an LLM generates a SOAP note S. Hallucination evaluation proceeds by:

\begin{figure}[h]
    \centering
    \includegraphics[width=0.8\columnwidth]{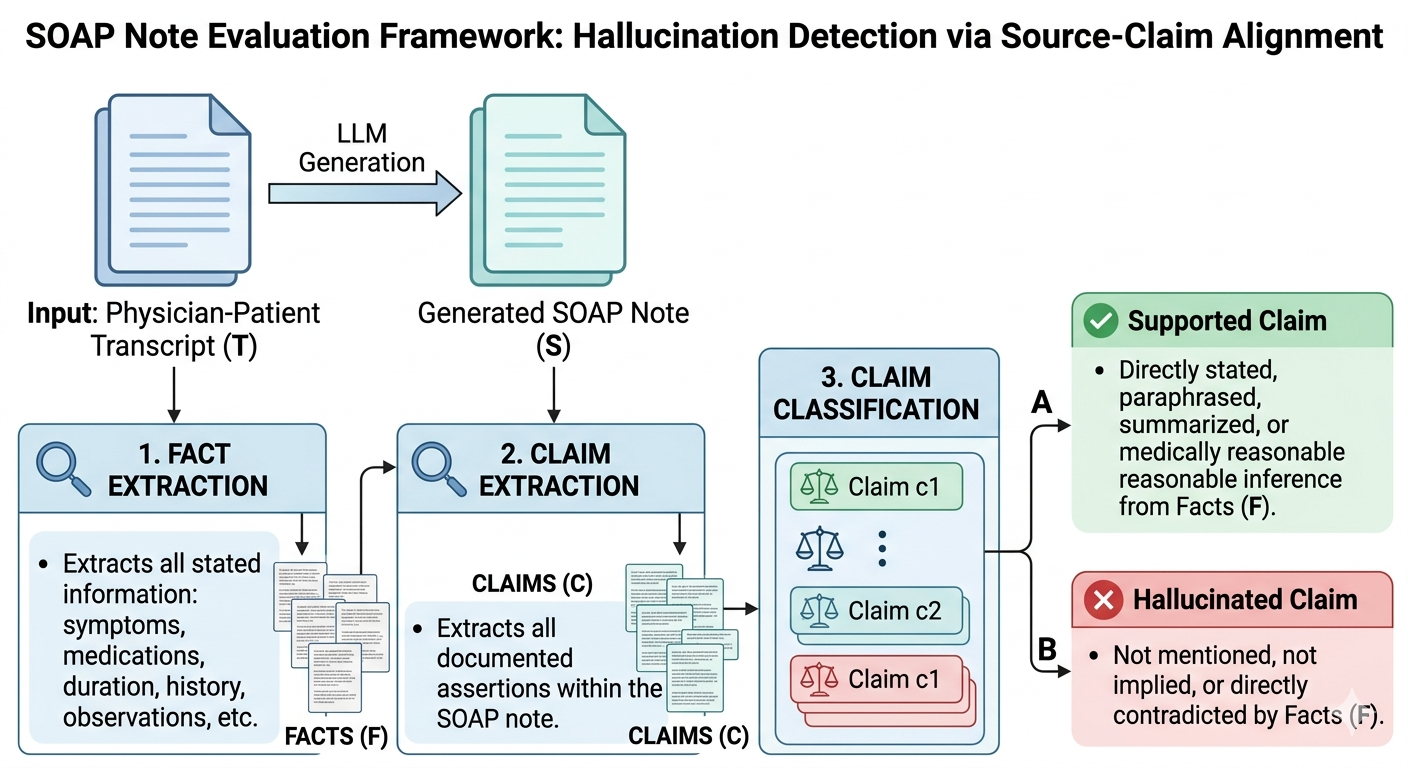}
    \caption{Source Claim Framework}
    \label{fig:your_label}
\end{figure}
\begin{enumerate}
    \item \textbf{Fact extraction:} A set of facts 
    $F = \{f_1, f_2, \ldots, f_n\}$ is extracted from $T$, 
    representing all stated information (symptoms, medications, duration, patient reported history, physician observations, etc.).

    \item \textbf{Claim extraction:} A set of claims 
    $C = \{c_1, c_2, \ldots, c_m\}$ is extracted from $S$, 
    representing all documented assertions in the SOAP note.

    \item \textbf{Claim classification:} Each claim $c_i$ is classified as:
    \begin{itemize}
        \item \textbf{Supported:} Directly stated, paraphrased, summarized, or a medically reasonable inference from $F$.
        \item \textbf{Hallucinated:} Not mentioned, not implied, or directly contradicted by $F$.
    \end{itemize}
\end{enumerate}

\subsection{The Inference Problem}
The critical challenge lies in the boundary between “supported by inference” and “hallucinated”. We propose a five tier classification in Table~\ref{tab:table_example}:
\begin{table}[ht]
\centering
\caption{Compact Claim Classification Tiers}
\label{tab:table_example}
\small
\setlength{\tabcolsep}{3pt}
\renewcommand{\arraystretch}{1.1}

\begin{tabularx}{\columnwidth}{|c|p{3.5cm}|X|c|}
\hline
\textbf{Tier} & \textbf{Type} & \textbf{Example} & \textbf{Verdict} \\
\hline
1 & Direct Statement & Patient says ``\textbf{I have a headache}''; SOAP: \textbf{headache} & Supported \\
\hline
2a & Paraphrasing & Patient reports ``\textbf{pain after every meal}''; SOAP: \textbf{postprandial pain} & Supported \\
\hline
2b & Trade name / generic name equivalence & Facts list: \textbf{Glucophage}; Claim: \textbf{Metformin} & Supported \\
\hline
3 & Inference & Patient describes ``\textbf{burning epigastric pain and antacid use}''; SOAP: \textbf{Likely PUD} & Supported \\
\hline
4 & Speculated Overreach & Patient states ``\textbf{no family history}''; SOAP: \textbf{family history of diabetes} & Hallucinated \\
\hline
5 & Contradiction & Patient ``\textbf{denies alcohol}''; SOAP: \textbf{alcohol related cause} & Hallucinated \\
\hline
\end{tabularx}
\end{table}

The distinction between Tier 3 and Tier 4 is where naive hallucination detectors fail. Tier 3 represents the kind of clinical reasoning that defines medical expertise pattern recognition from symptom clusters to diagnostic hypotheses. Tier 4 introduces information with no inferential basis whatsoever, whereas Tier 5 contradicts the information explicitly present in the conversation.

\subsection{Why This Distinction Matters}
Collapsing Tiers 3 and 4 into a single “hallucination” category has two serious consequences:
\begin{itemize}
    \item \textbf{Overcorrection risk:} If LLMs are penalized for valid clinical inference, fine tuning or prompting strategies may suppress medically necessary reasoning, producing documentation that is technically ``faithful'' to the transcript but clinically incomplete and potentially dangerous.
    
    \item \textbf{Evaluation invalidity:} Inflated hallucination scores misrepresent model performance, leading to incorrect conclusions about model suitability for clinical deployment.
\end{itemize}

\section{Methodology: LLM as Judge Framework}
\subsection{Data}
We constructed an evaluation dataset comprising 100 anonymized physician patient transcripts, covering a range of specialty including Cardiology, Dermatology, Endocrinology, ENT, Family Medicine etc. For each transcript, SOAP notes were generated using LLM.
\subsection{Stage 1: Inference Unaware Judge}
The initial LLM as Judge prompt was designed with strict factual grounding criteria. The judge was instructed to:
\begin{itemize}
    \item Extract all factual claims from the SOAP note.
    \item Cross reference each claim against the transcript.
     \item Flag any claim not explicitly stated in the transcript as hallucinated.
\end{itemize}
\vspace{0.5em}  

\textbf{Representative prompt excerpt (Stage 1):} \textit{"You are a medical fact checker. For each claim in the SOAP note, determine whether it is explicitly supported by the patient physician conversation transcript. If the claim contains information that was not directly stated in the conversation, mark it as HALLUCINATED."} \\ \\ 
Under this framework, the judge systematically flagged clinically valid inferences differential diagnoses, medical terminology translations of lay symptoms, and standard of care implications as hallucinations, resulting in a mean hallucination rate, well above clinically meaningful thresholds.

\subsection{Stage 2: Inference Aware Judge}

Stage 2 addresses the limitations of Stage 1 through two complementary components. The first is a restructured prompt that redefines the judge's reasoning criteria, detailed in the subsections below. The second is a retrieval augmented knowledge layer: for each claim under evaluation, relevant entries are retrieved from a curated clinical knowledge base and provided as additional context to the judge. This knowledge base is constructed from openly available medical sources, including SNOMED CT for clinical terminology and concept relationships, ICD-10 for diagnostic classifications, and supplementary drug knowledge resources covering generic and trade name equivalences. The retrieval step ensures that the judge's clinical reasoning is grounded not only in the model's parametric knowledge but in structured, standardized medical knowledge enabling more reliable synonym resolution, ontology aware inference, and terminology normalization than prompt design alone can achieve.

\subsubsection*{Supported Claim Criteria}
A claim is considered \textbf{supported}, and must not be flagged,  
under any of the following conditions:

\begin{itemize}
    \item The claim is a \textbf{paraphrase or terminology upgrade} of 
    a stated fact (e.g., ``throat irritation'' $\rightarrow$ 
    ``pharyngeal sensitivity'' $\rightarrow$ \textsc{supported}).

    \item The claim is a \textbf{medically reasonable inference} from 
    symptoms, history, or findings,  a diagnosis need not be 
    explicitly named by the physician to be valid.

    \item The claim reflects a \textbf{plan goal or intention} 
    consistent with a described treatment action.

    \item The claim is a \textbf{standard clinical summary statement} 
    appropriate given the overall picture (e.g., ``hemodynamically 
    stable'', ``in no acute distress'' $\rightarrow$ \textsc{supported} 
    if vitals are normal).
\end{itemize}

\subsubsection*{Clinical Knowledge Retrieval}
To support the reasoning criteria described above, the judge operates in a retrieval-augmented setting. Each claim is used as a query against a clinical knowledge base to surface relevant terminological and ontological context  for instance, confirming that two drug names refer to the same compound, that a symptom cluster maps to a recognized diagnostic category, or that a plan entry reflects a guideline-consistent intervention. This retrieved context is appended to the judge's input alongside the transcript facts and SOAP claims, allowing the model to draw on structured clinical knowledge when its parametric knowledge may be incomplete or inconsistent. The knowledge base draws on SNOMED CT, ICD-10, and open-source pharmacological resources, and was assembled to support precisely the categories of clinical inference most likely to be misclassified under a literal evaluation regime.

\subsubsection*{Synonym Rule (Non Negotiable)}
Trade names and generic drug names are treated as equivalent under all 
circumstances. If facts contain a trade name and the claim uses the 
generic, or vice versa, the claim is \textsc{supported}. The judge is instructed to treat trade names and generic drug names as equivalent under all circumstances, with the clinical knowledge base providing terminological grounding for such mappings at inference time.

\subsubsection*{Cross Section Verification}
Every claim must be verified against facts from all four SOAP sections, not only its own. A claim is supported if grounding 
evidence exists anywhere in the fact base. Stage 1 lacked this 
mechanism entirely ,  a Plan section claim referencing a penicillin 
allergy contraindication would be flagged as unsupported even when 
the allergy fact was present in Subjective.

\subsubsection*{Five Step Chain of Thought}
The judge applies the following protocol to every claim individually, 
with hallucination as the verdict of last resort:

\begin{enumerate}
    \item State the claim.
    \item Scan all four fact sections for a direct mention, synonym, 
    translated equivalent, or paraphrase.
    \item If found $\rightarrow$ \textsc{supported}. Stop.
    \item If not found, ask: is this a reasonable inference, standard 
    summary, or terminology upgrade? If yes $\rightarrow$ 
    \textsc{supported}. Stop.
    \item Otherwise $\rightarrow$ \textsc{hallucinated}. 
    State the specific reason.
\end{enumerate}

\subsubsection*{Retained Hallucination Conditions}
Stage 2 preserves a hard safety floor. A claim \textit{must} be 
marked hallucinated if it states a specific medication, dose, 
frequency, or duration with no equivalent in the facts; a diagnosis 
neither stated nor inferable from the symptom picture; a procedure 
or test not mentioned; or any detail that directly contradicts a 
stated fact. These conditions correspond to the Tier 4--5 fabrications 
in Table~\ref{tab:table_example} and represent the true patient safety 
boundary the framework enforces.

\textbf{Representative prompt excerpt (Stage 2):} \textit{"You are a clinical documentation expert and medical AI evaluator. When assessing SOAP note claims against the source transcript, apply the following criteria: A claim is SUPPORTED if it is (a) directly stated in the transcript, (b) a paraphrase or medical terminology equivalent of stated information, or (c) a medically reasonable inference that a trained clinician would draw from the presented symptom picture. A claim is HALLUCINATED only if it (a) introduces information with no basis in the transcript and cannot be reasonably inferred, or (b) directly contradicts information stated in the transcript."} \\ \\ 
\begin{table*}[p]
\centering
\caption{Comparison of claim assessments between Stage 1 (inference unaware) and Stage 2 
(inference aware). Tiers follow the classification in Table~\ref{tab:table_example}.}
\label{tab:tier_classi}
\small
\setlength{\tabcolsep}{4pt}
\renewcommand{\arraystretch}{1.3}
\begin{tabular}{|c|p{4cm}|p{4cm}|c|c|p{5cm}|}
\hline
\textbf{Tier} & \textbf{Claim} & \textbf{Stage 1 reasoning} & \textbf{S1} & \textbf{S2} & \textbf{Why they differ} \\
\hline
\multicolumn{6}{|l|}{\cellcolor{gray!12}\textbf{Tier 1 ,  direct restatement (both stages agree: supported)}} \\
\hline
1 & Penicillin allergy ,  rash
  & Verbatim match
  & \checkmark & \checkmark
  & No disagreement \\
1 & Ibuprofen 400\,mg daily $\times$ 6 months without food
  & Directly stated by patient
  & \checkmark & \checkmark
  & No disagreement \\
1 & Epigastric tenderness; no guarding; normal bowel sounds
  & Verbatim from physician exam
  & \checkmark & \checkmark
  & No disagreement \\
\hline
\multicolumn{6}{|l|}{\cellcolor{gray!12}\textbf{Tier 2a,  paraphrase / medical translation (Stage 1 flags, Stage 2 accepts)}} \\
\hline
2 & ``Nausea without emesis''
  & ``Emesis'' absent from transcript
  & \ding{55} & \checkmark
  & \textit{emesis} = \textit{vomiting}; medical synonym, no new information \\
2 & ``Postprandial and nocturnal pain''
  & Neither term in transcript
  & \ding{55} & \checkmark
  & Direct medical translation of ``after I eat'' and ``at night'' \\
2 & ``Retrosternal radiation''
  & ``Retrosternal'' absent
  & \ding{55} & \checkmark
  & Anatomical translation of ``goes up into my chest'' \\
2 & ``Patient denies dysphagia, melena, hematochezia''
  & Medical terms absent verbatim
  & \ding{55} & \checkmark
  & Clinical equivalents of patient's lay negations \\
\hline
\multicolumn{6}{|l|}{\cellcolor{gray!12}\textbf{Tier 2b ,  trade name / generic equivalence (Stage 1 flags, Stage 2 accepts)}} \\
\hline
2 & ``Patient is on lisinopril''
  & Facts list ``zestril''; ``lisinopril'' absent
  & \ding{55} & \checkmark
  & Synonym rule: zestril $\equiv$ lisinopril; same drug, different name\\
\hline
\multicolumn{6}{|l|}{\cellcolor{gray!12}\textbf{Tier 3 ,  clinical inference (Stage 1 flags, Stage 2 accepts)}} \\
\hline
3 & Diagnosis: Peptic Ulcer Disease (NSAID induced)
  & Physician never said ``PUD''
  & \ding{55} & \checkmark
  & Valid inference: NSAID use + symptom cluster + family history + H.~pylori order \\
3 & Diagnosis: GERD
  & ``GERD'' never named
  & \ding{55} & \checkmark
  & Valid differential from retrosternal burning + nocturnal symptoms \\
3 & ``Non penicillin H.~pylori regimen given allergy''
  & Antibiotic rationale not stated; allergy in Subjective, claim in Plan
  & \ding{55} & \checkmark
  & Cross section verification + safety inference: amoxicillin is standard therapy; allergy mandates alternative \\
3 & ``Alert, in no acute distress''
  & Phrase absent from transcript
  & \ding{55} & \checkmark
  & Standard documentation summary for ambulatory patient with normal vitals \\
3 & ``Consider GI referral if symptoms persist''
  & Not mentioned by physician
  & \ding{55} & \checkmark
  & Standard of care (ACG/NICE) for PUD/GERD unresponsive to empiric PPI \\
\hline
\multicolumn{6}{|l|}{\cellcolor{gray!12}\textbf{Tier 4 and 5,  genuine hallucination (both stages agree: unsupported)}} \\
\hline
4 & ``Family history of colorectal cancer''
  & No basis in transcript
  & \ding{55} & \ding{55}
  & Fabricated comorbidity; no inferential path from any stated fact \\
5 & ``Patient allergic to sulfonamides''
  & Contradicts transcript (penicillin only)
  & \ding{55} & \ding{55}
  & Direct contradiction of stated allergy history \\
\hline
\end{tabular}
\end{table*}

\section{Results}
We evaluate the two judge configurations Stage 1 (inference 
unaware) and Stage 2 (inference aware) across 100 
physician patient transcripts, measuring hallucination rates and comparing these against human annotator judgments. 
Table~\ref{tab:halluc_rates} summarizes the aggregate hallucination 
rates across the three evaluation conditions, and Figure~\ref{fig:three_graphs} 
visualizes the per sample distributions.

The results demonstrate a substantial reduction in flagged 
hallucinations when the judge is equipped with clinically informed 
criteria. Under Stage 1, the mean hallucination rate reached 35.2\%, 
nearly three times the human annotator baseline of 10.4\%. Stage 2 
reduced this to 9.1\%, achieving near parity with human judgment and 
confirming that the excess flagging under Stage~1 reflects evaluation 
design artifacts rather than genuine model errors.

\subsection{Hallucination Rate Comparison}
As reported in Table~\ref{tab:halluc_rates}, the gap between Stage~1 
and human annotators (24.8 percentage points) represents claims that 
trained clinicians considered valid but a strictly literal framework 
flagged as hallucinations. Stage~2 closes this gap to within 
1.3 percentage points, consistent with expected inter annotator 
variability in clinical review.

Table~\ref{tab:tier_classi} provides a granular claim level 
breakdown for a representative transcript (full transcript and 
generated SOAP note in Appendix~\ref{sec:appendix}), organized by 
the five tier classification of Section~\ref{sec:section_3}. 
Tier~2a false positives  medical translations of lay language 
such as ``postprandial pain'' for ``pain after I eat''  and 
Tier~2b errors, where Stage~1 penalized the generic name 
``lisinopril'' against the trade name ``zestril'' in the 
transcript, were fully resolved by Stage~2's synonym rule and 
paraphrase criteria. The most consequential disagreements occur at 
Tier~3: Stage~1 flagged clinically grounded diagnoses such as 
Peptic Ulcer Disease and GERD, and failed to link the penicillin 
allergy in the Subjective section to the non penicillin H.~pylori 
regimen in the Plan  a cross-section inference with direct 
patient safety implications. Stage~2's chain of thought protocol 
correctly reclassified all such cases as supported. Both stages 
agree on Tier~4--5 cases, confirming that inference awareness 
does not introduce permissiveness toward genuine fabrications or 
contradictions.

\begin{figure}[ht]
    \centering
    
    \begin{subfigure}[b]{0.3\textwidth}
        \centering
        \includegraphics[width=\textwidth]{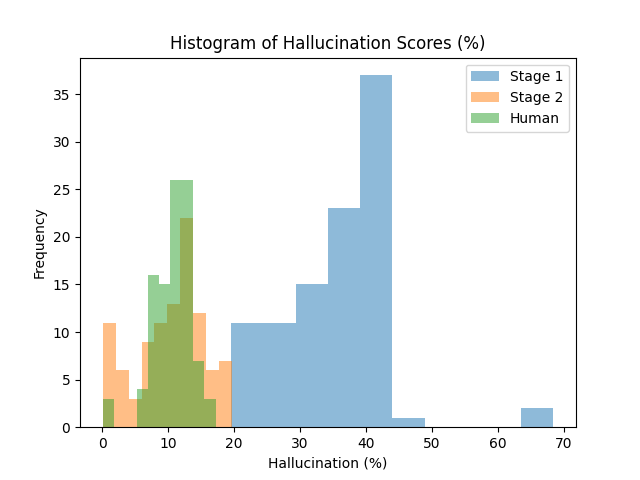}
        \caption{Plot 1}
    \end{subfigure}
    \hfill
    \begin{subfigure}[b]{0.3\textwidth}
        \centering
        \includegraphics[width=\textwidth]{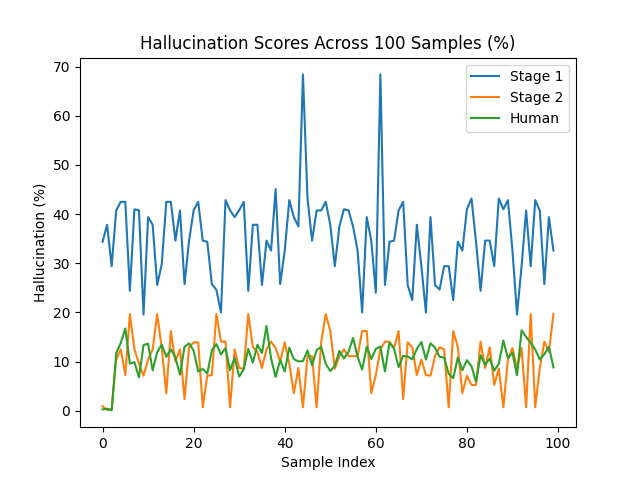}
        \caption{Plot 2}
    \end{subfigure}
    \hfill
    \begin{subfigure}[b]{0.3\textwidth}
        \centering
        \includegraphics[width=\textwidth]{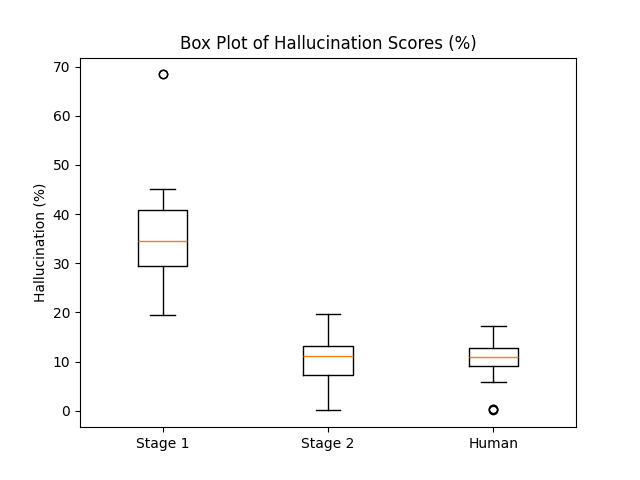}
        \caption{Plot 3}
    \end{subfigure}
    
    \caption{Visualization of hallucination scores across evaluation stages and human annotators.}
    \label{fig:three_graphs}
\end{figure}

Figure~\ref{fig:three_graphs} further illustrates these patterns 
across all 100 samples via three complementary visualizations.

\begin{table}[ht]
\centering
\caption{Comparison of hallucination rates across evaluation stages and human annotators}
\label{tab:halluc_rates}
\small
\renewcommand{\arraystretch}{1.2}
\begin{tabular}{|c|c|c|c|}
\hline
\textbf{Metric} & 
\makecell{\textbf{Stage 1} \\ (Inference Unaware)} & 
\makecell{\textbf{Stage 2} \\ (Inference Aware)} & 
\makecell{\textbf{Human Annotators}} \\
\hline
Mean Hallucination Rate & 35.2\% & 9.1\% & 10.4\% \\
\hline
\end{tabular}
\end{table}
\subsection{Analysis of Stage 1 False Positives}

Analysis of Stage 1 false positives revealed three dominant patterns of over flagging:

\begin{itemize}
    \item \textbf{Diagnostic inference:} Assessment section diagnoses (e.g., ``gastroesophageal reflux disease'' inferred from symptoms of heartburn and regurgitation) were consistently flagged despite strong symptom support.
    \item \textbf{Terminology translation:} Medical equivalents of lay descriptions (e.g., ``dyspnea on exertion'' for ``gets breathless when climbing stairs'') were flagged as unsupported.
    \item \textbf{Standard of care implications:} Plan section entries reflecting clinical guidelines (e.g., ordering an HbA1c for a patient with classic diabetes symptoms) were flagged despite clear inferential grounding.
\end{itemize}

Stage 2 correctly reclassified the vast majority of these as supported inferences, with residual hallucinations comprising primarily fabricated past medical history and unsupported medication entries.

\section{Conclusion}
The responsible deployment of LLMs in clinical documentation requires evaluation frameworks that reflect the realities of medical reasoning, not merely the mechanics of text matching. This paper demonstrates that naive hallucination detection  which flags any claim not explicitly stated in a source transcript dramatically overestimates hallucination rates in SOAP note generation, conflating clinically valid inference with fabrication.
By developing and validating an inference aware LLM as Judge evaluation framework which is grounded in both clinically calibrated prompt design and retrieval augmented medical knowledge, we show that hallucination rates can be assessed with greater clinical fidelity, yielding scores that align more closely with expert physician judgment. This matters not only for accurate model benchmarking, but for the clinical NLP community's ability to distinguish genuinely dangerous hallucination (fabricated medications, unsupported comorbidities, contradicted facts) from the inferential reasoning that is the very essence of clinical medicine.
As LLMs move from research prototypes to production clinical tools, the field must develop evaluation standards that are simultaneously rigorous in their demand for factual fidelity and sophisticated in their understanding of what clinical documentation actually requires. This work represents a step toward that goal.

\newpage
\section{Appendix}
\label{sec:appendix}

\begin{tcolorbox}[colback=gray!10, colframe=black, left=2pt, right=2pt, top=2pt, bottom=2pt, boxsep=2pt, arc=2pt,title=Sample Transcript]
\scriptsize
Doctor: Good morning, Mr. [ENTITY1]. What brings you in today? \\
Patient: Morning, doctor. I've been having this burning pain in my upper stomach right here in the middle for about the last three weeks. It's been pretty bad. \\
Doctor: Can you describe the pain more? Does it happen at any particular time? \\
Patient: It's mostly after I eat. And at night sometimes wakes me up. The burning goes up into my chest too. I've been popping antacids like candy but they only help for a little bit. \\
Doctor: Any nausea or vomiting? \\
Patient: Yeah, some nausea. No vomiting though. \\
Doctor: Any difficulty swallowing? \\
Patient: No, swallowing is fine. \\
Doctor: Have you noticed any blood in your stool or vomit, or have your stools been black or tarry? \\
Patient: No, nothing like that. Normal stools. \\
Doctor: Any weight loss lately? \\
Patient: Not intentionally. Maybe a couple pounds probably because eating makes the pain worse so I've been avoiding it. \\
Doctor: How long have you been on ibuprofen? It says here you've been taking it for your knee. \\
Patient: About six months for my knee pain. I take it pretty much every day. 400 mg. \\
Doctor: And you didn't always take it with food? \\
Patient: Sometimes, but not always. \\
Doctor: Any alcohol? \\
Patient: A glass of wine most evenings. \\
Doctor: Any smoking? \\
Patient: I quit about ten years ago. Smoked for about fifteen years before that. \\
Doctor: Any significant medical history? We have hypertension in your chart. \\
Patient: Yes, I've been on blood pressure medication for that Zestril. And my dad had stomach problems, ulcers I believe. \\
Doctor: Any known drug allergies? \\
Patient: Penicillin I got a rash from it years ago. \\
Doctor: Let me do a quick exam. You have some tenderness when I press here in the epigastric region. No guarding, no rigidity. Bowel sounds are normal. No palpable masses. \\
Patient: Is it serious? \\
Doctor: We'll need to do some tests. I want to check for H. pylori that's a bacteria that can cause ulcers. I'll order a urea breath test. I'm also going to order a complete blood count to check for anemia, and a comprehensive metabolic panel. I'm going to stop the ibuprofen for now and start you on something to protect your stomach lining a proton pump inhibitor. We'll go from there. \\
Patient: Okay, sounds good. \\
Doctor: Also, try to take it easy on the wine even moderate alcohol can irritate the stomach. Eat smaller, more frequent meals. Avoid spicy foods, coffee, and lying down right after eating. Follow up in four weeks, or sooner if things get worse. \\
Patient: Got it. Thank you, [ENTITY2].
\end{tcolorbox}

\lstdefinestyle{jsonstyle}{
  basicstyle=\scriptsize\ttfamily, 
  breaklines=true,                 
  columns=flexible,
  frame=none,
  showstringspaces=false,
  keywordstyle=\color{blue},
  stringstyle=\color{red},
  commentstyle=\color{green!50!black},
}

\begin{tcolorbox}[colback=gray!10, colframe=black,
                  left=2pt, right=2pt, top=2pt, bottom=2pt,
                  boxsep=2pt, arc=2pt,
                  title=Sample SOAP Note,
                  breakable]  
\begin{lstlisting}[style=jsonstyle]
{
  "Subjective": "54-year-old male presenting with a 3-week history of burning epigastric pain localized to the mid-epigastric region, worsening after meals and at night, frequently awakening him from sleep. Pain radiates retrosternally and is associated with nausea without emesis, heartburn, and reduced oral intake due to pain with eating, resulting in approximately 2 lbs of unintentional weight loss. Denies dysphagia, hematemesis, melena, or hematochezia. Has been self-medicating with over-the-counter antacids with only temporary relief. Reports daily use of ibuprofen 400 mg for approximately 6 months for chronic knee pain, often without food. Consumes approximately one glass of wine nightly. Former smoker with a 15 pack-year history, quit 10 years ago. Past medical history includes hypertension managed pharmacologically and chronic knee pain. Family history is significant for peptic ulcer disease. Allergic to penicillin. Current medications include lisinopril and ibuprofen 400 mg.",
  "Objective": "Alert, in no acute distress. Epigastric tenderness on palpation. No guarding or rigidity. Bowel sounds normal. No palpable masses.",
  "Assessment": "Peptic ulcer disease; rule out H. pylori co-infection; gastroesophageal reflux disease (GERD); hypertension.",
  "Plan": "Discontinue ibuprofen immediately. Initiate proton pump inhibitor. Order urea breath test, complete blood count, and comprehensive metabolic panel. Advise dietary modifications including small frequent meals and avoidance of irritants. Recommend reduction of alcohol intake. Monitor blood pressure. Reassess in 4-8 weeks and follow up in 4 weeks or sooner if symptoms worsen. Await H. pylori test results and treat if positive using a non-penicillin regimen. Consider gastroenterology referral and upper endoscopy if symptoms persist or alarm features develop."
}
\end{lstlisting}
\end{tcolorbox}

\bibliographystyle{unsrt}  
\nocite{Joshi2020}
\nocite{Abacha2023}
\bibliography{references}
\
\end{document}